\documentclass[runningheads]{llncs}

\usepackage[camera-ready,year=2026,ID=580]{eccv}

\usepackage{eccvabbrv}

\usepackage{graphicx}
\usepackage{booktabs}

\usepackage[accsupp]{axessibility}  
\usepackage{makecell}
\usepackage{colortbl}
\usepackage{multirow}
\usepackage{mathrsfs}

\usepackage{hyperref}
\usepackage[capitalize]{cleveref}

\usepackage{orcidlink}

\makeatletter
\def\daggerfoot{\gdef\@thefnmark{$\dagger$}\@footnotetext}
\makeatother

\begin{document}
\newcommand{\method}{{\fontfamily{lmtt}\selectfont\textbf{CogSENet}}\xspace}
\title{\method: Blind Image Deblurring with Blur-Conditioned Semantic Routing and Explicit Frequency Fusion}

\titlerunning{\method}

\author{
Pan Wang\inst{1}\orcidlink{0009-0003-4150-2920}\thanks{Equal contribution.}
\and
Yihao Hu\inst{2}\orcidlink{0009-0009-2566-1511}\textsuperscript{*, $\dagger$}
\and
Xiujin Liu\inst{3}\orcidlink{0009-0000-7008-1270}\textsuperscript{*}
}

\authorrunning{P.~Wang et al.}

\institute{
University of Science and Technology of China, Hefei, China\\
\email{diogenescask@gmail.com}
\and
Westlake University, Hangzhou, China\\
\email{mistletoehyh@gmail.com}
\and
University of Michigan, Ann Arbor, MI, USA\\
\email{jeanliu@umich.edu}
}

\maketitle
\daggerfoot{Corresponding author.}

\begin{abstract}

Blind image deblurring demands the recovery of high-fidelity details and coherent structures from complex, unknown degradations. Current blind image deblurring methods struggle with real-world, spatially varying degradations, and lack the semantic awareness necessary to reliably differentiate valid textures from artifacts. To bridge this gap, we propose \method, a dynamic, semantic-aligned reconstruction framework inspired by the eagle's visual system. By mimicking the eagle’s active saccadic scanning, we devise a Semantic-Driven State Space Module (SDSSM) with semantic-aware token regrouping via differentiable routing, enabling prompt-conditioned long-range dependency modeling. To ensure physically interpretable recovery of textures and structures, a BiFreqFusionBlock (BFFB) mirrors functional differentiation of the eagle's retina by decomposing features into high and low frequencies using wavelet transforms. Finally, we estimate a continuous Blur Field (CBF) from blur image and fuse it with CLIP semantic priors to modulate the deepest latent features, emulating focal adaptation and enabling adaptive restoration under spatially non-uniform blur. Extensive experiments demonstrate that \method outperforms state-of-the-art deblurring methods in both visual quality and structural fidelity with fewer parameters, while also performing favorably on dehazing, deraining, and denoising tasks.

\end{abstract}

\section{Introduction}

Blind image deblurring (BID) is a fundamental yet challenging low-level vision task, aiming to recover high-fidelity details and coherent structures from images degraded by unknown and complex blur. Unlike non-blind settings where blur kernels are known, BID requires the network to implicitly or explicitly estimate the degradation while reconstructing the clear image. The field has evolved significantly, shifting from Convolutional Neural Networks (CNNs)\cite{nafnet}, which prioritize local extraction but miss global context, to Vision Transformers\cite{restormer}, which capture long-range dependencies at high computational costs. Recently, State Space Models (SSMs)\cite{mambair,vmamba},  have emerged as a powerful alternative, offering linear complexity and effective global modeling. 

\label{sec:intro}
\begin{figure}[!t]
  \centering
  \includegraphics[width=0.9\linewidth]{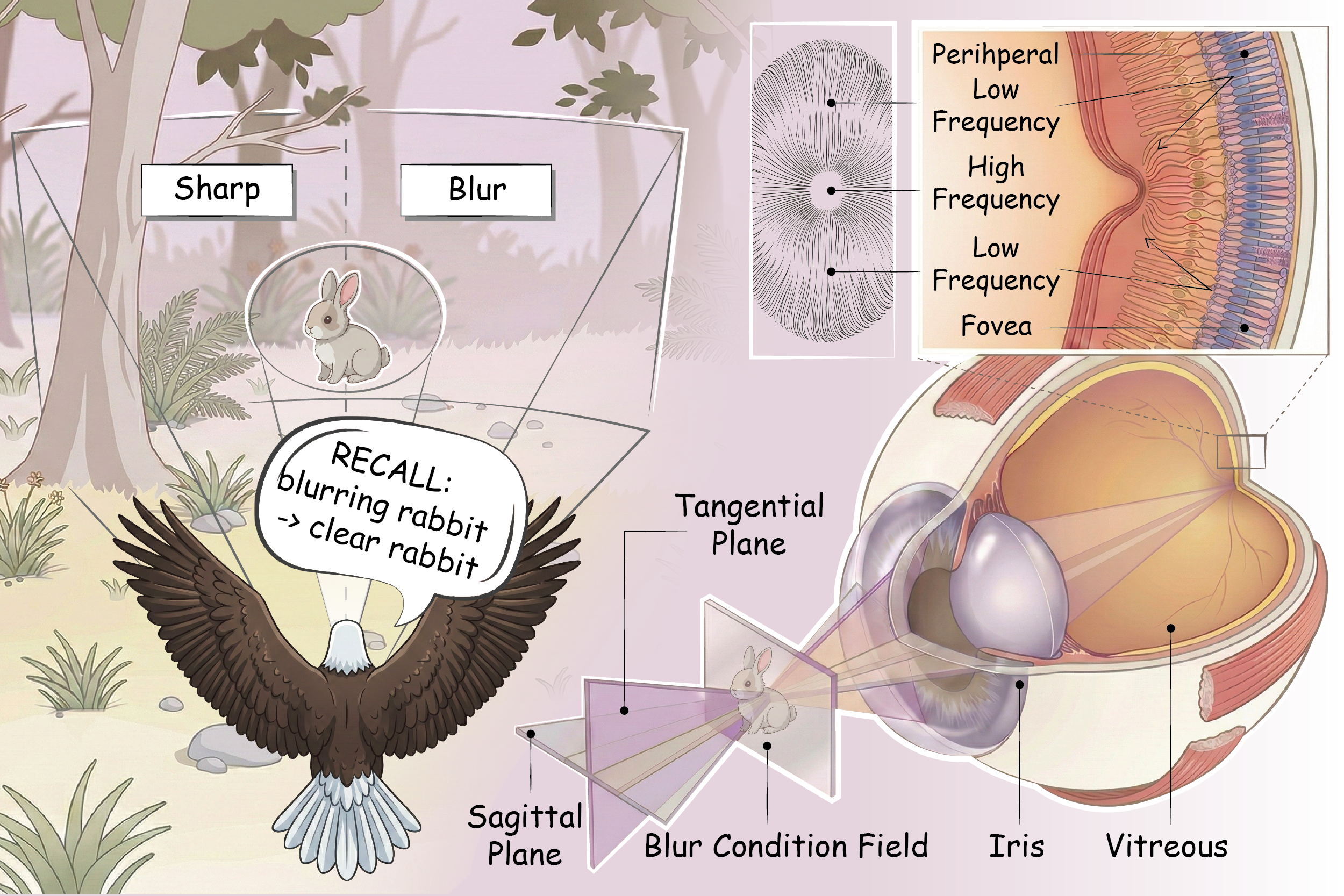}
  \caption{
  Biological inspiration of \method. Emulating the eagle's visual system: (1) Active saccadic scanning for semantic targets inspires our SDSSM to perform semantic-grouped scanning; (2) Retinal differentiation motivates our BFFB to decouple high/low frequencies; (3) Continuous focal adaptation inspires our Continuous Blur Field to model spatially varying blur; (4) Memory recall inspires a frozen CLIP prior to provide high-level semantic guidance at the deepest features.
  }
  \label{fig:eagle inspire}
\end{figure}

However, a fundamental gap persists: most existing methods treat restoration as a static pixel-mapping regression. They often overlook the intrinsic link between high-level semantic understanding and low-level physical degradation. Real-world degradations are inherently non-uniform and spatially varying. This gap becomes more pronounced under complex real-world blurs. Current restoration methods fail because they process image features indiscriminately, lacking mechanisms to explicitly model semantic groupings or separate structural frequencies. Standard SSMs and frequency filters often process image content indiscriminately, failing to distinguish semantic boundaries or decouple complex textures from structural frequencies. Moreover, by relying on static, discrete estimations, these models cannot capture the continuous, spatially-varying trajectory of actual degradation. Thus, distinguishing valid textures from artifacts requires the model to have semantic awareness. This motivates our shift from static regression to a dynamic, semantically aligned reconstruction process, where the network understands \emph{what} it is restoring to guide \emph{how} it restores it.

To bridge the gap, as shown in \cref{fig:eagle inspire}, we draw inspiration from the highly adaptive visual system of an eagle, which maintains extraordinary clarity across diverse environments through four specialized mechanisms. In this work, we reframe blind image restoration from a static pixel-mapping regression to a dynamic, semantic-aligned reconstruction process, and propose \method, a unified framework that elevates blind deblurring from a passive pixel-regression to an active, semantic-aligned reconstruction. By shifting the focus to semantic-driven state modeling and explicit frequency fusion, we aim to bridge the gap between semantic perception and physical texture recovery, enabling the network to understand \emph{what} to restore based on semantic context and \emph{how} to restore it via interpretable frequency analysis. To address these challenges, \method harnesses the eagle's natural biological strengths through three core modules working in synergy:

First, mirroring the eagle’s continuous focal adaptation, we estimate a Continuous Blur Field (CBF) and inject it into the deepest latent features before bottleneck processing. Rather than relying on discrete kernel assumptions, CBF spatially varying degradation in a dense form. It is then jointly used with frozen CLIP semantic priors to modulate deep features in a content-adaptive manner, improving restoration under complex non-uniform motion blur. Second, to emulate the functional differentiation of the eagle's retina, we design BiFreqFusionBlocks (BFFB). By explicitly disentangling features into high-acuity foveal textures and broad-field structural cues via wavelet transforms, BFFB processes these components through specialized transformer and dense convolution branches, ensuring a physically interpretable recovery. Finally, we internalize the eagle’s active saccadic scanning via Semantic-Driven State Space Modules (SDSSM). Instead of uniform 2D selective scan, SDSSM performs differentiable hard routing into semantic groups and scans the semantically regrouped sequence. This design enables a prompt-conditioned, non-causal scanning mechanism to capture the global semantic dependencies beyond fixed causal order.

The main contributions of this paper are summarized as follows:
\begin{itemize}
    \item We transform long-range dependency modeling from uniform scanning to semantic-driven routing by non-causal token allocation within Semantic-Driven State Space Modules (SDSSM), enabling prompt-modulated selective scan over semantically regrouped tokens.
    \item We reframe feature processing from implicit spatial convolution to explicit frequency disentanglement by employing wavelet-based decomposition and specialized high/low frequency branches (BFFB), ensuring the physically interpretable recovery of textures and structures.
    \item We elevate degradation modeling from discrete kernel prediction to continuous blur-field perception, and jointly leverage frozen CLIP semantic priors for deep feature modulation, providing robust conditioning for complex non-uniform blur.
\end{itemize}

Through this bio-mimetic integration, \method fundamentally departs from prior restoration paradigms by unifying semantic routing, frequency-aware reconstruction, and continuous degradation conditioning. Extensive experiments show that, compared with existing SSM-based methods, our approach restores blind images with fewer parameters while outperforming state-of-the-art methods.

\section{Related works}
\label{sec:related}
\noindent \textbf{CNN-based image restoration methods.}  
In recent years, convolutional neural networks (CNNs) have been the dominant paradigm for low-level image restoration due to their strong locality bias and efficient computation~\cite{dong2014learning,dong2015compression,zhang2017beyond}. Early approaches typically built encoder--decoder or multi-scale architectures to enlarge receptive fields, with later designs introducing residual learning and attention to improve texture recovery. While these CNN-based models achieve strong performance across various tasks, their structural rigidity limits their capacity to handle spatially varying degradations where global context is essential. To overcome this local bias, our framework introduces a CLIP-conditioned Continuous Blur Field (CBF). By explicitly injecting global semantic priors, our approach enables continuous, spatially adaptive degradation modeling far beyond the reach of conventional discrete convolutions.

\noindent \textbf{Transformer-based restoration methods.} 
To better model global interactions, Transformers adapt self-attention to dense prediction, capturing long-range correlations that are difficult for purely convolutional designs~\cite{conde2022swin2sr,swinir}. To make attention practical for high-resolution inputs, works often adopt windowed, hierarchical, or hybrid CNN--Transformer blocks. Despite their effectiveness, attention-based global modeling incurs high computational costs and can indiscriminately propagate noise or erroneous details when the degradation is ill-posed. In contrast, our Semantic-Driven State Space Module (SDSSM) replaces quadratic self-attention with non-causal dynamic routing. This ensures that only contextually relevant tokens interact, effectively suppressing the propagation of artifacts while maintaining linear complexity.
 
\noindent \textbf{State-space models for image restoration.} 
More recently, state-space models (SSMs) have emerged as an alternative for global modeling with linear complexity~\cite{gu2021efficiently,smith2022simplified,fu2022hungry}. By using structured recurrent-style propagation (e.g., selective scanning~\cite{huang2024localmamba}), SSMs offer a favorable trade-off between receptive field and computational efficiency. Nevertheless, without explicit control, global propagation in SSMs can diffuse errors across distant regions and amplify high-frequency artifacts. Furthermore, existing constrained SSMs largely treat image features as homogeneous signals. To address this unexplored bottleneck, our Bi-Frequency Fusion Block (BFFB) explicitly disentangles high-frequency foveal textures from broad low-frequency structures via wavelet transforms. This explicit frequency decoupling ensures physically interpretable and stable state-space propagation, setting our method apart from existing generic SSM designs.

\section{Method}

\begin{figure}[tb]
  \centering
  \includegraphics[width=\linewidth]{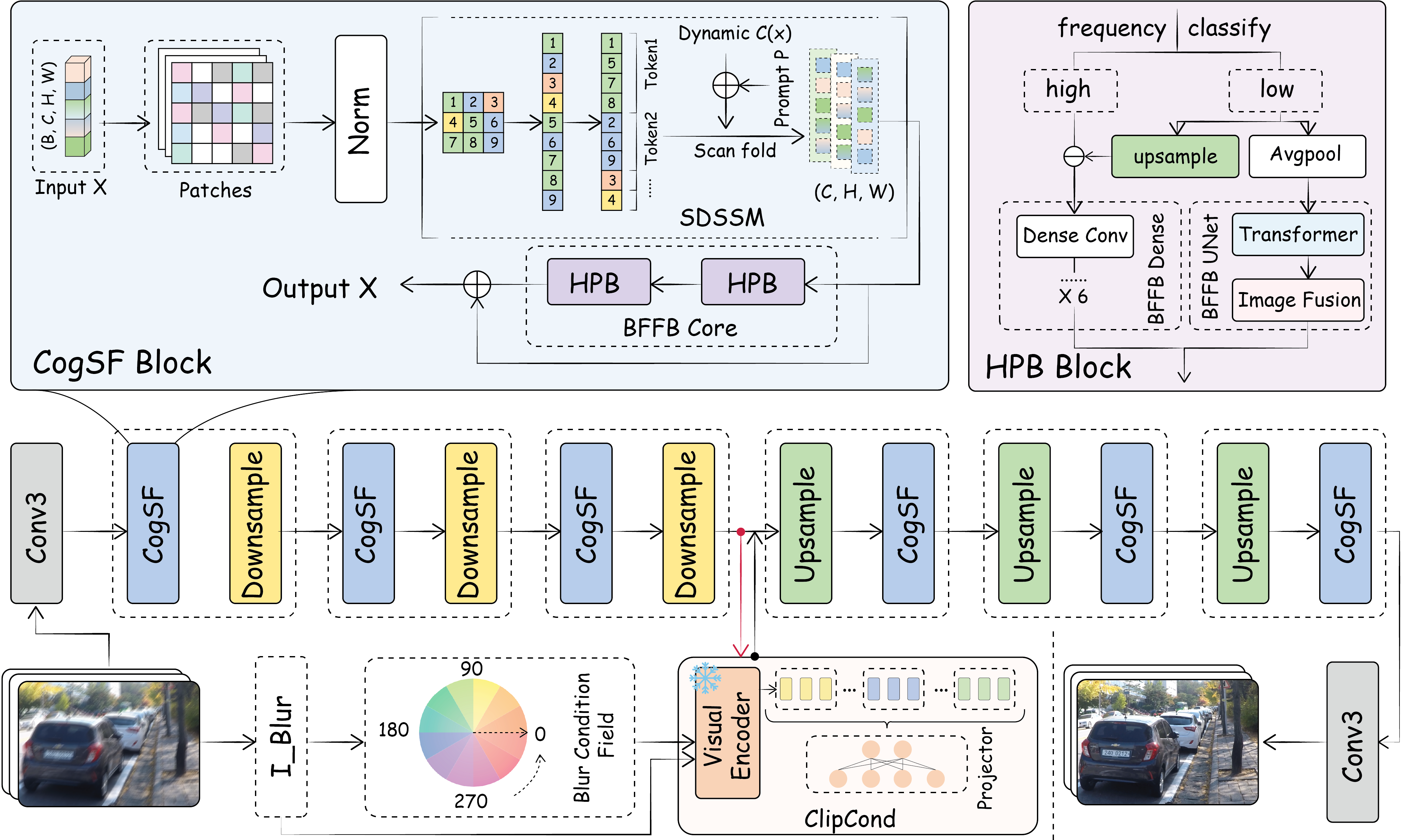}
  \caption{
  Our proposed \method. A U-shaped backbone is built with stacked CogSF blocks across multi-scale encoder-decoder stages. Each CogSF block integrates an SDSSM module for prompt-guided dynamic token routing and a BFFB core composed of HPB units for explicit high/low-frequency decoupling and fusion. At the deepest bottleneck, we inject a continuous blur field conditioned on frozen CLIP embeddings to modulate bottleneck features.
  }
  \label{fig:model}
\end{figure}

\subsection{Overall Architecture}
Our goal is to develop a semantically guided and frequency-aware restoration network that recovers sharp structures while remaining computationally practical. To this end, we propose \method, a hierarchical encoder-decoder architecture built upon stacked CogSF blocks.

As shown in \cref{fig:model}, given a degraded image $\mathbf{I}_{in}\in\mathbb{R}^{H\times W\times 3}$, we first apply a $3\times3$ overlap convolution to extract shallow features $\mathbf{F}_s\in\mathbb{R}^{H\times W\times C}$, where $C$ is the base channel width. The shallow feature is then fed into a three-level symmetric encoder-decoder backbone. At level $l$ ($l\in\{1,2,3\}$), features have spatial resolution $\frac{H}{2^{l-1}}\times\frac{W}{2^{l-1}}$ and channel dimension $2^{l-1}C$. Each level consists of multiple CogSF blocks. A CogSF block is composed of two complementary modules: an SDSSM module for semantic context aggregation and a BFFB module for explicit bi-frequency refinement (see Secs.~\ref{sec:cogsf}).

For transitions across scales, we use bilinear resizing for both downsampling and upsampling, followed by a $3\times3$ convolution. Specifically, downsampling reduces the spatial resolution by half while doubling the number of channels, whereas upsampling restores the resolution and correspondingly halves the channel dimension. We further introduce lateral skip connections between encoder and decoder features at the same scale to preserve fine spatial details.

Different from a standard U-shaped design, we inject top-down guidance only at the bottleneck. After the level-3 encoder, we predict a continuous blur field, implemented as a dense 2D displacement field, from the input using a lightweight convolutional head. We embed the blur field and fuse it with the bottleneck feature map via a lightweight feature fusion layer, producing blur-aware descriptors.

Meanwhile, a frozen CLIP image encoder extracts a global semantic embedding from the input. We then compute a spatial attention map by cosine matching between the semantic embedding and the blur-aware descriptors, and use this attention to modulate the \emph{original} bottleneck features with a single learnable scalar. The resulting bottleneck representation, enriched with both semantic cues and blur awareness, is then decoded and progressively integrated with higher-resolution skip connections, enabling a coarse-to-fine restoration process.

Finally, a lightweight $3\times3$ reconstruction head predicts a residual image, which is added to the input via an identity shortcut to stabilize optimization and preserve low-frequency content.

\subsection{CogSF Blocks}
\label{sec:cogsf}
CogSF is the basic building block of \method. Each CogSF sequentially applies a Semantic-Driven State-Space Module (SDSSM) to aggregate long-range context in a structured manner, followed by a Bi-Frequency Fusion Block (BFFB) to explicitly refine complementary frequency components.In practice, BFFB has two instantiations: FullBFFB, which follows Eq.~(4), and LightBFFB, an efficient approximation using average-pooling-based high/low splitting and lightweight implicit frequency branches.

\noindent \textbf{Semantic-Driven State-Space Module (SDSSM)}
\label{sec:sdssm}
Generic self-attention lacks explicit semantic organization, while small-kernel convolutions are strictly local. Because degradations are spatially non-uniform, SDSSM couples Gumbel-Softmax semantic routing with efficient regrouping for selective-scan state-space modeling, enabling interaction among distant but semantically related regions.

\paragraph{Semantic routing and prompt construction.}
Given a feature $\mathbf{X}\in\mathbb{R}^{B\times C\times H\times W}$ flattened into tokens $\mathbf{T}=\{\mathbf{t}_i\}_{i=1}^L \in\mathbb{R}^{L\times C}$, we sample an assignment vector $\mathbf{p}_i \in \mathbb{R}^K$ over $K$ clusters via Gumbel-Softmax relaxation. Concurrently, we compute its semantic prompt $\mathbf{z}_i \in \mathbb{R}^{d_s}$ by taking the expectation over a learnable codebook $\mathcal{E} \in \mathbb{R}^{K \times d_s}$:
\begin{equation}
p_{i,k}=\frac{\exp((\mathbf{w}_{\theta,k}^\top \mathbf{t}_i + g_k)/\tau)}{\sum_{j=1}^K \exp((\mathbf{w}_{\theta,j}^\top \mathbf{t}_i + g_j)/\tau)}, \quad \mathbf{z}_i=\sum_{k=1}^K p_{i,k} \mathbf{e}_k
\end{equation}
where $g_k \sim \mathrm{Gumbel}(0,1)$ and $\tau$ is the temperature. This generates a prompt sequence $\mathbf{Z}=\{\mathbf{z}_i\}_{i=1}^L$. The continuous $\mathbf{p}_i$ preserves end-to-end differentiability during backpropagation.

\paragraph{Semantic regrouping and prompt-modulated selective scan.}
We derive a permutation operator $\mathcal{P}_\pi$ by sorting tokens based on their dominant cluster indices ($k^*_i = \arg\max_k p_{i,k}$). To prevent gradient truncation from the non-differentiable argmax, we explicitly detach the sorting indices and route gradients backward exclusively through the continuous prompt $\mathbf{Z}$. We apply $\mathcal{P}_\pi$ to form semantically contiguous sequences: $\tilde{\mathbf{T}}=\mathcal{P}_\pi(\mathbf{T})$ and $\tilde{\mathbf{Z}}=\mathcal{P}_\pi(\mathbf{Z})$. Scanning along this semantic sequence—combined with alternating spatial flips across network blocks—allows the native 1D causal scan to inherently achieve non-causal global perception. For the regrouped sequence $\tilde{\mathbf{t}}_j \in \mathbb{R}^{d_{inner}}$, the scan operates via discretized matrices $\bar{\mathbf{A}}_j$ and $\bar{\mathbf{B}}_j$. We inject context-aware dynamics by adding $\tilde{\mathbf{z}}_j$ to the emission matrix $\mathbf{C}_j$:
\begin{equation}
\mathbf{h}_j=\bar{\mathbf{A}}_j \mathbf{h}_{j-1} + \bar{\mathbf{B}}_j \tilde{\mathbf{t}}_j, \quad \tilde{\mathbf{y}}_j=(\mathbf{C}_j + \tilde{\mathbf{z}}_j)\mathbf{h}_j + \mathbf{D}\tilde{\mathbf{t}}_j
\end{equation}
Here, $\mathbf{h}_j \in \mathbb{R}^{d_s}$. The state parameters $\bar{\mathbf{B}}_j \in \mathbb{R}^{d_s}$, $\mathbf{C}_j \in \mathbb{R}^{d_s}$, and the step $\Delta_j$ (used for $\bar{\mathbf{A}}_j$) are dynamically projected from $\tilde{\mathbf{t}}_j$. To maintain efficiency, these projection weights and the skip parameter $\mathbf{D} \in \mathbb{R}^{d_{inner}}$ are strictly shared across all semantic groups.

\paragraph{Output restoration.}
Finally, we apply the inverse permutation $\mathcal{P}_\pi^{-1}$ and sequence-to-grid folding $\mathcal{R}(\cdot)$ to form the residual update:
\begin{equation}\mathbf{X}_{\text{out}}=\mathbf{X} + \mathcal{R}(\mathcal{P}_\pi^{-1}(\tilde{\mathbf{Y}}))\end{equation}

\noindent \textbf{Bi-Frequency Fusion Block (BFFB)}
\label{sec:bffb}
While SDSSM provides global context, high-quality restoration also requires explicitly handling frequency components: low-frequency structures govern global appearance, whereas high-frequency details determine sharpness. BFFB addresses this by combining an explicit high/low decomposition pathway via wavelet transforms with an efficient spectral refinement branch, adaptively fusing their outputs.

\paragraph{Explicit high/low modeling.}
To achieve physically interpretable frequency disentanglement, we employ the orthogonal wavelet projection $\mathcal{W}$ (e.g., Haar wavelet) to decompose the input feature map $\mathbf{X}$ into low-frequency approximation coefficients $\mathbf{X}_{LL}$ and high-frequency detail coefficients $\mathbf{X}_{H}$. The explicit high/low modeling branch, denoted as $\mathcal{H}(\cdot)$, processes these decomposed sub-bands independently. Specifically, a low-frequency branch $\Phi_{low}$ models global structures using a lightweight U-shaped module, while a high-frequency branch $\Phi_{high}$ enhances local details using dense convolutions. 

\paragraph{Frequency-domain refinement.}
In parallel, the frequency-domain refinement branch $\mathcal{F}(\cdot)$ maps the signal to the Fourier domain for adaptive spectral filtering. By reweighting the frequency responses via a learned spectral filter $\mathbf{M}$ and projecting back to the spatial domain, this branch explicitly suppresses artifacts and recovers fine textures. This operation leverages the discrete Fourier transform $\mathscr{F}$ and its inverse $\mathscr{F}^{-1}$ to ensure rigorous spectral modulation.

\paragraph{Adaptive fusion.}
The final restored feature $\mathbf{X}_{\text{out}}$ is obtained via an adaptive, parameterized residual fusion that balances the physical structure from the wavelet domain and the spectral texture from the Fourier domain:
\begin{equation}
\mathbf{X}_{\text{out}}=\mathbf{X} + \beta\,\mathcal{W}^{-1}\big( \Phi_{low}(\mathbf{X}_{LL}), \Phi_{high}(\mathbf{X}_{H}) \big) + (1-\beta)\,\mathscr{F}^{-1}\big( \mathbf{M} \odot \mathscr{F}(\mathbf{X}) \big),
\end{equation}
where $\mathcal{W}^{-1}$ is the inverse wavelet transform, $\odot$ denotes element-wise multiplication, and $\beta$ is a learnable scalar dynamically adapting to the content to weight the two branches. This design enables content-aware restoration that jointly preserves robust global structures and sharp fine details.

\subsection{Continuous Blur Field and Semantic Joint Modulation}
\label{sec:cbf_clip}

To emulate focal adaptation under spatially non-uniform degradations, we propose to jointly modulate the deepest latent features using both physical and high-level cognitive priors. As shown in our network bottleneck, this is achieved through a Continuous Blur Field (CBF) coupled with CLIP semantics.

Concretely, a lightweight CNN extracts a degradation prior from the input image $\mathbf{I}_{in}$ and outputs a continuous flow field $\mathbf{u}\in\mathbb{R}^{B\times 2\times H\times W}$. To stabilize optimization and constrain blur magnitude, we bound the field with $\tanh(\cdot)$ and scale it by a maximum displacement $d_{\max}$:
\begin{equation}
\mathbf{u} = d_{\max} \cdot \tanh\big(\mathrm{CNN}_{blur}(\mathbf{I}_{in})\big).
\end{equation}

We apply CBF only at the deepest bottleneck to inject degradation awareness into global-context features. Given bottleneck features $\mathbf{F}\in\mathbb{R}^{B\times C\times h\times w}$, we first resize $\mathbf{u}$ to the bottleneck resolution and encode it with a convolutional embedding. The encoded degradation feature is then concatenated with $\mathbf{F}$ and fused by a lightweight $1\times1$ convolution to obtain blur-aware descriptors $\tilde{\mathbf{F}}$.

In parallel, a frozen CLIP image encoder extracts a global semantic token $\mathbf{c}\in\mathbb{R}^{d}$. To make cosine similarity well-defined across modalities, we map CLIP semantics and blur-aware descriptors into the same channel dimension before $\ell_2$ normalization. Then, for each sample, we compute a spatial attention map by pixel-wise cosine similarity followed by spatial softmax:
\begin{equation}
A^{(b)}_{i,j}=\frac{\exp\big( \langle \bar{\mathbf{c}}^{(b)}, \bar{\tilde{\mathbf{F}}}^{(b)}_{i,j} \rangle \big)}{\sum_{x=1}^{h} \sum_{y=1}^{w} \exp\big( \langle \bar{\mathbf{c}}^{(b)}, \bar{\tilde{\mathbf{F}}}^{(b)}_{x,y} \rangle \big)},
\end{equation}
where $\langle \cdot, \cdot \rangle$ denotes inner product, and the bar notation represents $\ell_2$-normalized features.

Finally, we modulate the \emph{original} bottleneck feature map $\mathbf{F}$ using residual attention gating:
\begin{equation}
\mathbf{F}'_{i,j}=\mathbf{F}_{i,j} + \gamma \, (A_{i,j} \cdot \mathbf{F}_{i,j}),
\end{equation}
where $\gamma$ is a learnable scalar initialized to zero. This preserves identity mapping at early training and gradually introduces semantic-physical guidance as training proceeds. The guided representation $\mathbf{F}'$ is then fed into the decoder for coarse-to-fine restoration.

\section{Experiments}

\begin{table*}[t]
  \centering
  \renewcommand\arraystretch{1.1}
  \caption{Quantitative comparisons on GoPro, HIDE, RealBlur-R and RealBlur-J datasets. PSNR and SSIM are reported separately. \textbf{Boldface} and \underline{underlined} indicate the best and second-best results, respectively.}
  \label{tab:gopro_hide_realblur}
  \resizebox{\textwidth}{!}{
  \begin{tabular}{c|c|cc|cc|cc|cc}
    \toprule
    \multirow{2}{*}{\textbf{Methods}} 
    & \multirow{2}{*}{\makecell{\textbf{Params}\\(M)$\downarrow$}}
    & \multicolumn{2}{c|}{\textbf{GoPro}} 
    & \multicolumn{2}{c|}{\textbf{HIDE}}
    & \multicolumn{2}{c|}{\textbf{RealBlur-R}}
    & \multicolumn{2}{c}{\textbf{RealBlur-J}} \\
    \cline{3-10}
    & 
    & PSNR (dB)$\uparrow$ & SSIM$\uparrow$ 
    & PSNR (dB)$\uparrow$ & SSIM$\uparrow$
    & PSNR (dB)$\uparrow$ & SSIM$\uparrow$
    & PSNR (dB)$\uparrow$ & SSIM$\uparrow$ \\
    \midrule
    Restormer~\cite{restormer}
    & 26.1 & 33.57 & 0.9656 & 31.22 & 0.9420 & 36.19 & 0.9570 & 28.96 & 0.8790 \\
   
    MPRNet~\cite{mprnet}
    & 20.1 & 32.66 & 0.9590 & 30.96 & 0.9390 & 39.31 & 0.9720 & 31.76 & 0.9220 \\
    
    NAFNet~\cite{nafnet}
    & 67.9 & 33.71 & 0.9668 & 31.31 & 0.9427 & 39.80  & 0.9731  & 32.40 & 0.9249 \\
    
    ALGNet~\cite{algnet}
    & 54.6 & 33.49 & 0.9640 & 31.64 & 0.9470 & 36.35 & 0.961 & 29.12 & 0.8860 \\

    SRN~\cite{SRN}
    & \textbf{6.8} & 30.26 & 0.9342 & 28.36 & 0.9040 & 38.65 & 0.9652 & 31.38 & 0.9091 \\

    MIMO-Unet+~\cite{MIMOUnet}
    & 16.1  & 32.45 & 0.9567 & 29.99 & 0.9304 & 38.55 & 0.9650 & 31.92 & 0.9190 \\

    Stripformer~\cite{stripformer}
    & 19.7 & 33.08 & 0.9624 & 31.03 & 0.9395 & 39.84 & 0.9737 & 32.48 & 0.9290 \\

    AdaRevD~\cite{adarev}
    & 152.4 & 34.50 & 0.9710 & 32.26 & 0.9520 & 36.56 & 0.957 & 30.12 & 0.8940 \\

    EVSSM~\cite{evssm}
    & 17.1  & 34.51 & 0.9713 & 31.99 & 0.9503 & 41.27 & 0.9776 & 34.34 & 0.9456 \\
    
    MLWNet~\cite{MLWNet}
    & 24.1 & 33.8 & 0.9676 & 31.06 & 0.9320 & 40.69 & 0.9760 & 33.84 & 0.9410 \\

    BANet~\cite{banet}
    & 18.0 & 32.54 & 0.9570 & 30.16 & 0.9300  & 39.55 & 0.971 & 32.00 & 0.9230 \\

    MRLPFNet~\cite{MRLPFNet}
    & 20.6 & 33.50 & 0.9650 & 31.63 & 0.9465 & 40.92 & 0.9753 & 33.19 & 0.9361 \\

    BANet+~\cite{banet}
    & 40.0 & 33.03 & 0.9610 & 30.58 & 0.9350  & 39.90 & 0.972 & 32.42 & 0.9290 \\

    FFTformer~\cite{fftformer}
    & 16.6 & 34.21 & 0.9692 & 31.62 & 0.9455 & 40.11 & 0.9753 & 32.62 & 0.9326 \\
    \rowcolor[HTML]{E7E6E6}
    Ours
    & \underline{8.9} & \underline{34.72} & \underline{0.9744} & \underline{32.18} & \underline{0.9514} & \underline{41.83} & \underline{0.9799} & \underline{34.54} & \underline{0.9473} \\
    \rowcolor[HTML]{E7E6E6}
    Ours+
    & 19.4 & \textbf{34.91} & \textbf{0.9757} & \textbf{32.42} & \textbf{0.9533} & \textbf{41.91} & \textbf{0.9810} & \textbf{34.72} & \textbf{0.9488} \\
     
    \bottomrule
  \end{tabular}}
\end{table*}

\subsection{Experimental Settings}

\noindent \textbf{Dataset and Evaluation.}
Following recent state-of-the-art image restoration methods, we evaluate our approach on benchmark datasets spanning image deblurring, deraining, dehazing, and low-light enhancement. For image deblurring, we use the widely adopted GoPro\cite{gopro}, HIDE\cite{hide}, and RealBlur datasets\cite{realblur}. RealBlur contains two subsets, RealBlur-J and RealBlur-R, constructed with different post-processing pipelines, and comprises 3,758 training images and 980 testing images in total. For cross-task evaluation, we use Rain100L\cite{rain100} for deraining, RESIDE-SOTS\cite{reside} for dehazing, and BSD68\cite{bsd68} for denoising task. For fair comparison, we strictly follow the standard training and testing protocols of each dataset. Unless otherwise stated, quantitative performance is reported using PSNR and SSIM.

\noindent \textbf{Training Details.}
The shallow feature dimension is set to 48, and the numbers of backbone blocks at the three hierarchical levels are [6, 6, 12]. We train the model using AdamW and apply standard data augmentation, including random horizontal/vertical flips and rotations. Training follows a three-stage progressive strategy. In Stage 1, we train with a patch size of $128 \times 128$ and a total batch size of 32 for 300,000 iterations, with the learning rate annealed from $1 \times 10^{-3}$ to $1 \times 10^{-7}$. In Stage 2, the patch size is increased to $256 \times 256$ and the total batch size is reduced to 8 for another 300,000 iterations; the learning rate is reset to $5 \times 10^{-4}$ and annealed to $1 \times 10^{-7}$. In Stage 3, we further increase the patch size to $320 \times 320$ and reduce the total batch size to 4 for 150,000 iterations, initializing the learning rate at $1 \times 10^{-4}$ and gradually decaying it to $1 \times 10^{-7}$. We use cosine annealing for learning-rate scheduling throughout. For cross-task training, we disable the blur-field branch. In our implementation, this branch predicts a dense 2D continuous blur field from the input image and injects it into the deepest features via a conditional fusion module for blur-specific, motion-aware modulation. Because these blur-specific cues are not required for non-deblurring tasks, disabling this branch avoids introducing task-mismatched blur priors. Unless otherwise stated, all experiments are implemented in PyTorch and conducted on four NVIDIA L40 GPUs. The training code and models are available at \href{https://github.com/Genk641/CogSENet}{CogSENet}.

\begin{figure}[tb]
  \centering
  \includegraphics[width=\linewidth]{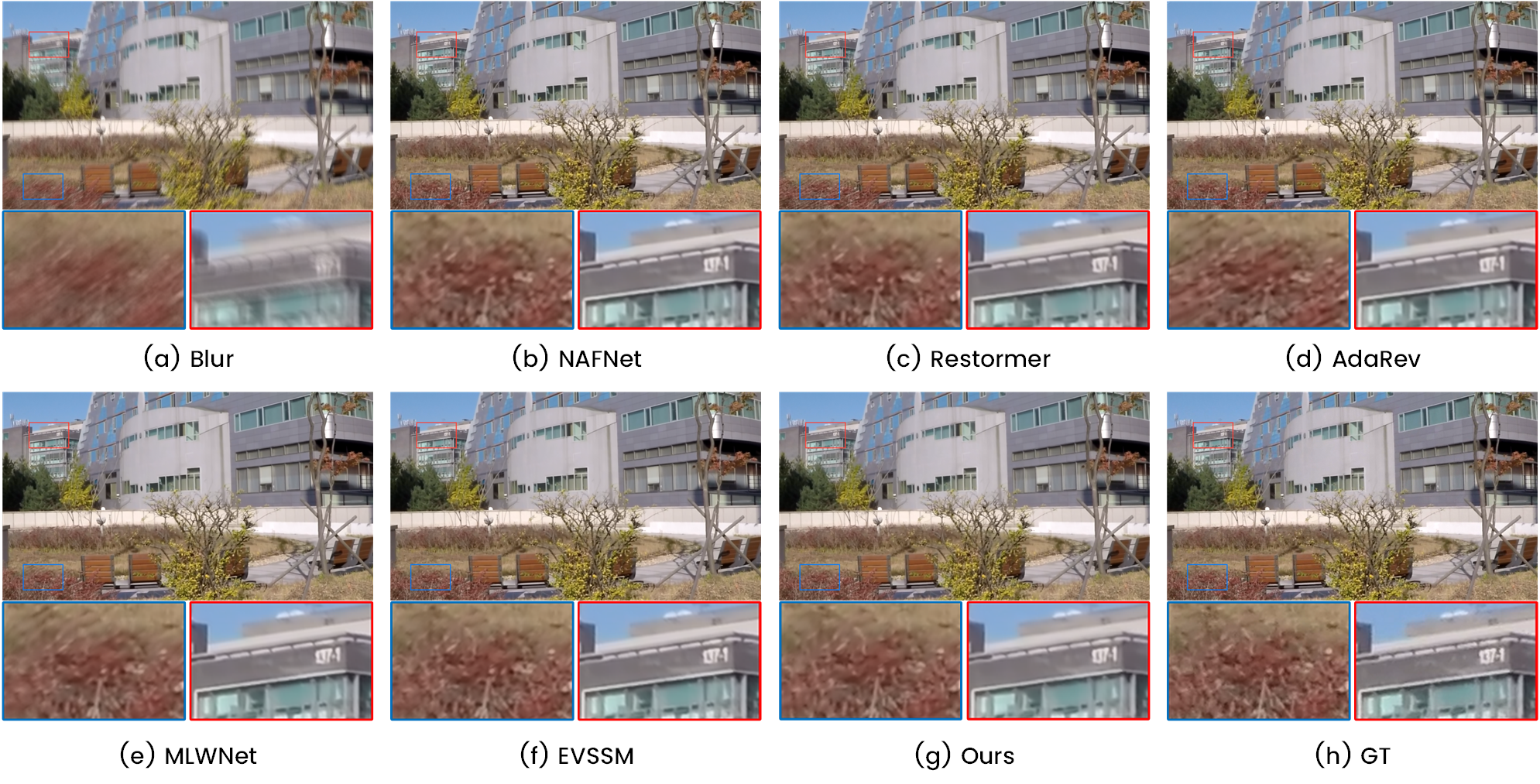}
  \caption{
  Visual comparison on the GoPro dataset. Compared with the methods in (b)–(f), the proposed method produces a deblurred image with clearer structures and finer details in (g), yielding results that are more faithful to the ground truth in (h).
  }
  \label{fig:gopro_result}
\end{figure}

\noindent \textbf{Blind Image Motion Deblurring.} Our denotes the lightweight configuration, which uses LightBFFB throughout the backbone for efficiency. Our+ denotes a hybrid configuration, which keeps LightBFFB in the backbone while introducing a FullBFFB at the bottleneck to enhance representational capacity. Furthermore, the CLIP model is strictly frozen and acts solely as an external semantic prior. Since it requires no gradient updates, its parameters are excluded from the total model complexity.

\cref{tab:gopro_hide_realblur} reports quantitative comparisons on GoPro, HIDE, RealBlur-R, and RealBlur-J.
Our base model achieves 34.72 dB PSNR and 0.9744 SSIM on GoPro with only 8.9M parameters, yielding a favorable accuracy-efficiency trade-off.
Compared with representative efficient methods, Our improves PSNR by 0.51 dB over FFTformer (34.21 dB) and by 0.21 dB over EVSSM (34.51 dB), while using substantially fewer parameters (8.9M vs.\ 16.6M/17.1M).
Building on this lightweight baseline, Our+ further improves restoration quality across all deblurring benchmarks, validating the benefit of stronger bottleneck frequency fusion.

For cross-dataset generalization, we follow the common protocol of training on GoPro and testing on HIDE.
Under this setting, Our reaches 32.18 dB PSNR and 0.9514 SSIM, and Our+ provides further gains under the same protocol.
On real-world benchmarks, Our obtains 41.83/0.9799 (PSNR/SSIM) on RealBlur-R and 34.54/0.9473 on RealBlur-J; Our+ remains consistently stronger, indicating improved robustness to diverse real blur patterns and imaging conditions.

\begin{figure}[tb]
  \centering
  \includegraphics[width=\linewidth]{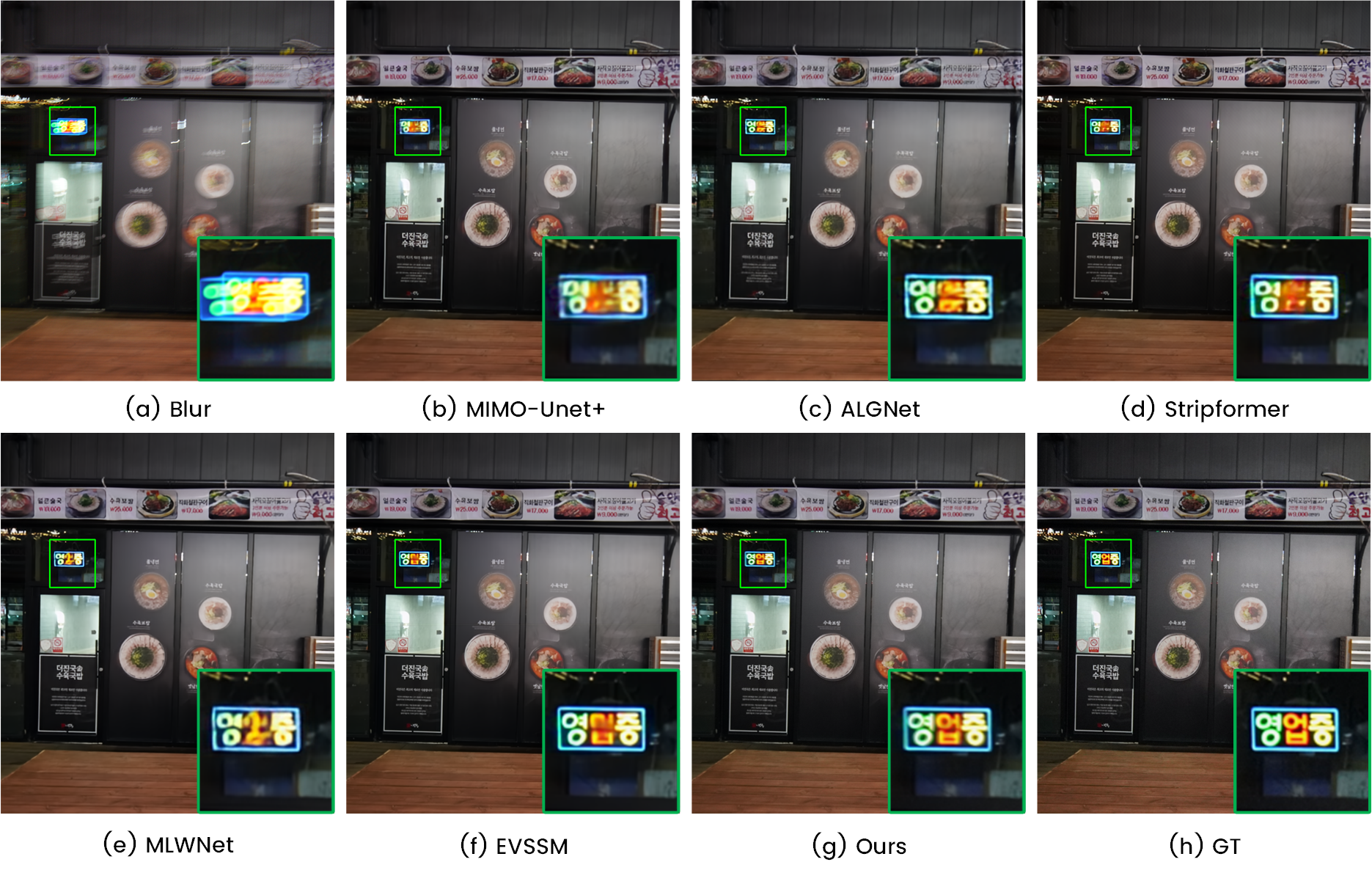}
  \caption{
  Deblurred results on the RealBlur dataset. The characters and structural details in (b)–(f) are still blurry, while our method restores sharper characters and clearer structural details.
  }
  \label{fig:realblur_result}
\end{figure}

Qualitative comparisons are shown in \cref{fig:gopro_result} and \cref{fig:realblur_result}. On GoPro, previous methods tend to leave residual blur around fine structures and high-frequency regions, while our method recovers cleaner edges and sharper local textures. On RealBlur, especially in character and contour regions, competing methods still produce over-smoothed details, whereas our reconstruction preserves clearer boundaries and more faithful structures. These visual observations are consistent with the quantitative improvements in \cref{tab:gopro_hide_realblur}.

\begin{table*}[t]
\centering
\caption{Quantitative comparison on the image deraining task.}
\label{tab:derain}
\resizebox{\textwidth}{!}{
\begin{tabular}{c|cc|cc|cc|cc|cc|cc|cc}
\toprule
\textbf{Dataset}
& \multicolumn{2}{c|}{\textbf{MSPFN}~\cite{mspfn}}
& \multicolumn{2}{c|}{\textbf{LPNet}~\cite{lpnet}}
& \multicolumn{2}{c|}{\textbf{MPRNet}~\cite{mprnet}}
& \multicolumn{2}{c|}{\textbf{Restormer}~\cite{restormer}}
& \multicolumn{2}{c|}{\textbf{SwinIR}~\cite{swinir}}
& \multicolumn{2}{c|}{\textbf{IDT}~\cite{idt}}
& \multicolumn{2}{c}{\textbf{Ours}} \\
Rain100
& PSNR & SSIM & PSNR & SSIM & PSNR & SSIM & PSNR & SSIM & PSNR & SSIM & PSNR & SSIM & PSNR & SSIM \\
\midrule
Rain100L
& 32.40 & 0.9330
& 34.26 & 0.9500
& 36.40 & 0.9650
& 38.99 & 0.9780
& \underline{39.02} & \underline{0.9788}
& \textbf{39.05} & \textbf{0.9792}
& \underline{39.02} & 0.9785 \\

Rain100H
& 28.66 & 0.8600
& 23.73 & 0.8100
& 30.41 & 0.8900
& 31.46 & 0.9040
& 31.75 & 0.9050
& \textbf{32.17} & \underline{0.9080}
& \underline{32.15} & \textbf{0.9099} \\
\bottomrule
\end{tabular}%
}
\end{table*}

\begin{table*}[t]
\centering
\caption{Quantitative comparison on image denoising and dehazing tasks.}
\label{tab:denoise_dehaze}
\resizebox{\textwidth}{!}{
\begin{tabular}{c|c|cc|cc|cc|cc|cc|cc|cc}
\toprule
\multirow[c]{2}{*}{\textbf{Task}} & \multirow[c]{2}{*}{\textbf{Dataset}}
& \multicolumn{2}{c|}{\textbf{BRDNet}~\cite{brdnet}}
& \multicolumn{2}{c|}{\textbf{LPNet}~\cite{lpnet}}
& \multicolumn{2}{c|}{\textbf{FDGAN}~\cite{fdgan}}
& \multicolumn{2}{c|}{\textbf{MPRNet}~\cite{mprnet}}
& \multicolumn{2}{c|}{\textbf{DL}~\cite{DL}}
& \multicolumn{2}{c|}{\textbf{AirNet}~\cite{airnet}}
& \multicolumn{2}{c}{\textbf{Ours}} \\
& 
& PSNR & SSIM & PSNR & SSIM & PSNR & SSIM & PSNR & SSIM & PSNR & SSIM & PSNR & SSIM & PSNR & SSIM \\
\midrule
\multirow{3}{*}{Denoise}
& BSD68 ($\sigma=15$)
& \underline{34.10} & 0.9291
& 32.31 & 0.9236
& 31.11 & 0.9147
& 34.01 & 0.9334
& 33.25 & 0.9225
& \textbf{34.14} & \textbf{0.9356}
& 34.02 & \underline{0.9336} \\

& BSD68 ($\sigma=25$)
& 31.43 & 0.8847
& 27.87 & 0.8674
& 29.57 & 0.8770
& 31.34 & 0.8892
& 30.38 & 0.8679
& \underline{31.48} & \underline{0.8928}
& \textbf{31.50} & \textbf{0.8931} \\

& BSD68 ($\sigma=50$)
& 28.16 & 0.7942
& 25.71 & 0.7656
& 27.12 & 0.7895
& 28.10 & 0.8014
& 26.68 & 0.7415
& \underline{28.23} & \underline{0.8057}
& \textbf{28.29} & \textbf{0.8062} \\
\midrule
Dehaze
& SOTS
& 23.31 & 0.9116
& 21.43 & 0.8631
& 23.15 & 0.9207
& \underline{28.21} & \underline{0.9672}
& 24.68 & 0.9243
& 23.18 & 0.9000
& \textbf{28.72} & \textbf{0.9692} \\
\bottomrule
\end{tabular}%
}
\end{table*}

\noindent \textbf{Single Image Deraining.}
\cref{tab:derain} reports the quantitative comparison on the Rain100L and Rain100H benchmarks. While most state-of-the-art methods saturate on the light rain dataset (Rain100L), our method demonstrates remarkable robustness on the significantly more challenging Rain100H dataset. Specifically, our model achieves the highest SSIM and highly competitive PSNR, trailing the best PSNR by merely 0.02 dB. Notably, compared to the prominent Restormer, our approach yields a substantial +0.69 dB PSNR improvement on Rain100H. This performance leap under severe degradation confirms that our frequency-aware design (BFFB) effectively decouples high-frequency dense rain streaks from the underlying background, while the SDSSM maintains global structural integrity.

\noindent \textbf{Single Image Denoising.}
As shown in \cref{tab:denoise_dehaze}, our framework delivers leading performance on the BSD68 dataset across varying noise levels. For mild noise ($\sigma=15$), our method performs comparably to top-tier approaches. More importantly, our architectural superiority becomes increasingly pronounced as the noise severity escalates. At higher noise levels ($\sigma=25$ and $\sigma=50$), our model consistently establishes the new state-of-the-art in both PSNR and SSIM. This trend indicates that explicitly modeling the high/low-frequency components allows the network to accurately isolate and suppress high-frequency noise without compromising true image textures in heavily corrupted regimes.

\noindent \textbf{Single Image Dehazing.}
\cref{tab:denoise_dehaze} further details the evaluation on the standard SOTS dehazing benchmark. Our model secures the top rank, achieving 28.72 dB PSNR and 0.9692 SSIM. Impressively, it outperforms the strong baseline MPRNet by a significant margin of +0.51 dB. Atmospheric haze typically causes global contrast degradation coupled with local detail attenuation. Our superior performance demonstrates that the content-adaptive routing in SDSSM effectively captures long-range dependencies for global contrast recovery, while BFFB refines the attenuated high-frequency details, leading to artifact-free dehazing.

\subsection{Analysis and Discussion}
We have shown that \method achieves a favorable accuracy--efficiency trade-off against prior deblurring methods. In this section, we present ablation studies to analyze the contributions of its key components. Unless otherwise specified, all ablation variants are trained on the GoPro dataset for 300K iterations using \(128 \times 128\) patches, the AdamW optimizer (initial learning rate \(1\times10^{-3}\), \(\beta_1=0.9\), \(\beta_2=0.99\), weight decay \(5\times10^{-4}\)), and a True Cosine Annealing schedule with 5K warm-up iterations and a minimum learning rate of \(1\times10^{-7}\). The total batch size is 64 on 4 GPUs.

\begin{figure}[!tb]
\centering
\scriptsize
\begin{minipage}[t]{0.47\linewidth}
    \centering
    \captionsetup{width=0.88\linewidth}
    \captionof{table}{SDSSM ablations on GoPro.}
    \label{tab:sdssm_ablation}
    \setlength{\tabcolsep}{1.2pt}
    \begin{tabular}{l|cccc}
    \toprule
    Method & Par. & Mem. & Time & PSNR \\
           & (M)  & (G)  & (ms) & /SSIM \\ \midrule
    w/o prompt  & 8.9 & 22 & 92.11 & 34.15/0.9685 \\
    w/o regroup & 8.9 & 22 & 92.19 & 34.21/0.9691 \\
    w/o SDSSM   & 9.1 & 22 & 92.54 & 33.95/0.9677 \\ \bottomrule
    \end{tabular}
\end{minipage}\hfill
\begin{minipage}[t]{0.47\linewidth}
    \centering
    \captionsetup{width=0.88\linewidth}
    \captionof{table}{BFFB ablations on GoPro.}
    \label{tab:bffb_ablation}
    \setlength{\tabcolsep}{1.2pt}
    \begin{tabular}{l|cccc}
    \toprule
    Method & Par. & Mem. & Time & PSNR \\
           & (M)  & (G)  & (ms) & /SSIM \\ \midrule
    w/o H/L Split & 8.7 & 20 & 92.17 & 34.01/0.9680 \\
    w/o Imp. Freq & 8.8 & 21 & 92.18 & 34.15/0.9689 \\
    w/o BFFB  & 7.9 & 20 & 90.02 & 33.92/0.9677 \\ \bottomrule
    \end{tabular}
\end{minipage}
\end{figure}

\noindent \textbf{Effects of SDSSM.}
Table~\ref{tab:sdssm_ablation} investigates SDSSM on the GoPro dataset. Removing SDSSM entirely degrades performance (33.95 dB) despite incurring higher parameter and latency costs, confirming the superior efficiency of our state-space design. Inside SDSSM, discarding the semantic prompt (34.15 dB) deprives state transitions of essential context-aware dynamics. Furthermore, omitting semantic regrouping (34.21 dB) restricts the scan to rigid spatial axes, failing to associate distant but semantically related regions. By coupling regrouping with prompt modulation, the full SDSSM achieves peak quality while maintaining highly favorable computational footprints.

\begin{figure}[tb]
  \centering
  \includegraphics[width=0.9\linewidth]{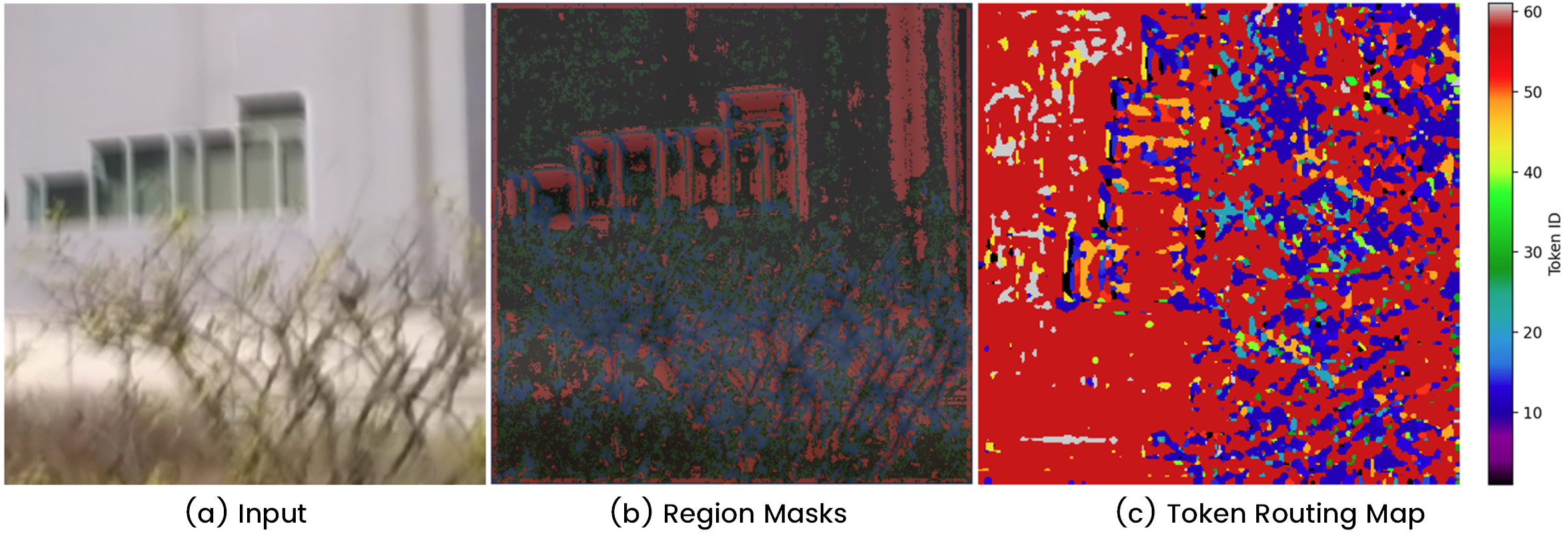}
  \caption{
  Region-aware token routing in SDSSM.
  }
  \label{fig:sdssm_effect}
\end{figure}

\cref{fig:sdssm_effect} further provides qualitative evidence.
The routing map is semantically consistent with the degradation masks: large smooth/motion-dominated areas are assigned to dominant token groups, while texture-rich regions (e.g., shrubs and structural edges) activate diverse token IDs. 
This behavior confirms that SDSSM learns content-adaptive specialization and allocates representation capacity to spatially varying blur patterns.

\noindent \textbf{Effects of BFFB.}
We ablate the frequency-aware designs within BFFB in \cref{tab:bffb_ablation}. Discarding the explicit high/low-frequency split severely degrades performance, confirming that decoupling low-frequency contents from high-frequency details is essential for accurate restoration. Furthermore, omitting the implicit frequency branch restricts the network's capacity for dynamic frequency modulation, dropping the PSNR to 34.15 dB. Removing BFFB entirely yields the lowest performance. Overall, BFFB delivers consistent and substantial quality improvements with only a trivial architectural footprint, proving its efficacy as a foundational component.

\begin{figure}[tb]
  \centering
  \includegraphics[width=\linewidth]{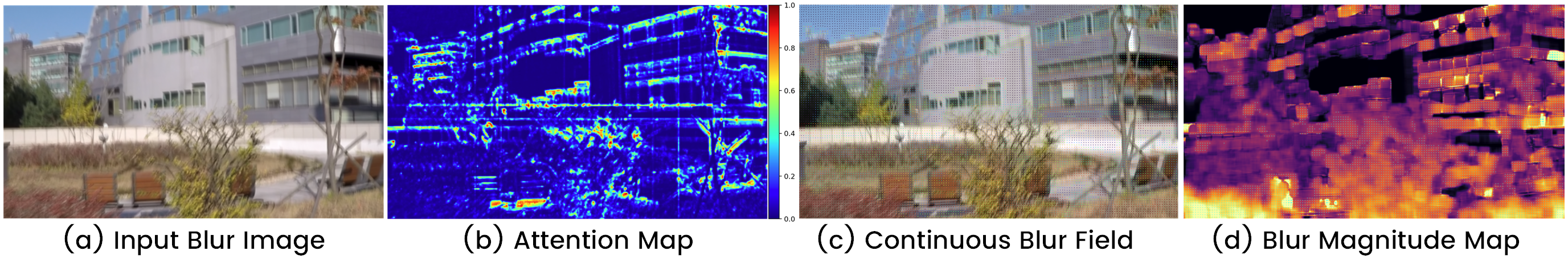}
  \caption{
  Visualization of Continuous Blur Field and Semantic Priors. (a) Input blurry image. (b) CLIP-guided attention map, highlighting semantically informative structures. (c) Estimated continuous blur field. (d) Corresponding blur magnitude map.
  }
  \label{fig:cbf_effect}
\end{figure}

\noindent \textbf{Effectiveness of Continuous Blur Field and Semantic Priors.}
As quantitatively verified in \cref{tab:ablation_cbf_clip}, applying either the physical blur field or cognitive CLIP semantics individually yields limited gains, whereas their joint modulation substantially boosts performance, proving them highly complementary. \cref{fig:cbf_effect} visualizes the behavior of our joint modulation module. First, the network directly estimates a spatially continuous blur field from the input to explicitly capture the non-uniform degradation. As shown in (c) and (d), the blur field and its magnitude map are structure-aligned, naturally assigning higher blur intensity to heavily degraded regions while preserving smooth transitions. 

\begin{table}[t]
\centering
\caption{Ablation on Continuous Blur Field and Semantic Priors.}
\label{tab:ablation_cbf_clip}
\renewcommand{\arraystretch}{0.85}
\setlength{\tabcolsep}{3pt}
\begin{tabular}{l|c|c|c|c}
\toprule
Dataset & \method (Ours) & w/ only CBF & w/ only CLIP & w/ only CBF\&CLIP \\
\midrule
GoPro & 34.31/0.9719 & 33.80/0.9668  & 33.78/0.9667 & 33.88/0.9673 \\
HIDE  & 31.68/0.9470 & 31.32/0.9431 & 31.29/0.9430 & 31.40/0.9433 \\
\bottomrule
\end{tabular}
\end{table}

By integrating the physical degradation prior with deep latent features and querying it via high-level CLIP semantics, we obtain the attention map visualized in (b). This semantic-driven modulation effectively isolates informative high-frequency structures while suppressing feature responses in non-informative flat regions. Consequently, our joint modulation emulates focal adaptation: it dynamically steers the network to prioritize high-fidelity restoration in severely degraded semantic regions, effectively averting the over-amplification of noise or artifacts in weakly blurred areas.

\begin{figure}[t]
  \centering
  \includegraphics[width=\columnwidth]{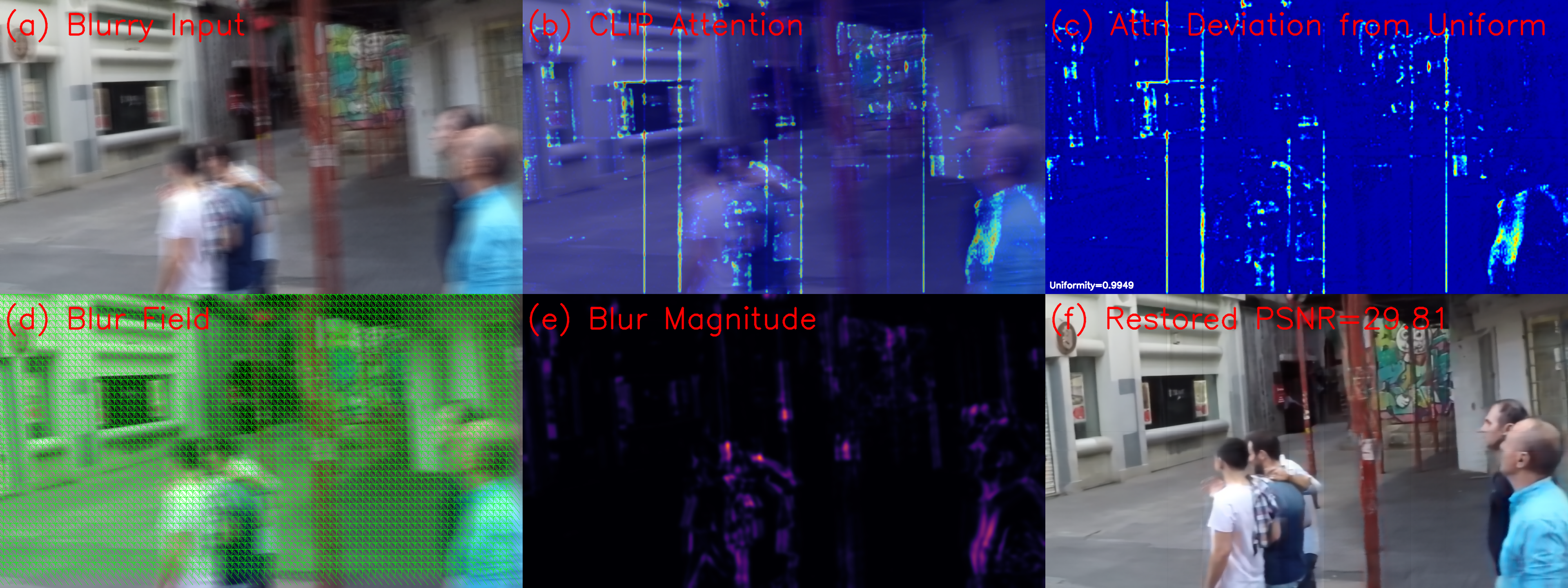}
  \label{fig:failue_vis}
  \caption{Failure analysis.}
\end{figure}

\noindent \textbf{Limitations.} Our physical-semantic joint modulation relies on frozen CLIP priors. As shown in \cref{fig:failue_vis}, our model occasionally struggles with extreme motion blur that destroys distinct semantic cues. In such failure cases, the CLIP encoder fails to extract valid semantics, causing our focal adaptation to degrade to nearly uniform attention (b-c) and the blur field to miss complex local motions (d-e), ultimately leaving residual blur and vertical artifacts in the restored image (f). Future work includes integrating degradation-robust semantic extractors and extending this architecture to video restoration.

\section{Conclusion}
In this paper, we propose an efficient region- and frequency-aware framework for image deblurring. We introduce the Semantic-Driven State-Space Module (SDSSM) for linear-complexity, content-adaptive long-range modeling, and the Bi-Frequency Fusion Block (BFFB) to explicitly decouple and refine high/low-frequency components. To tackle spatially non-uniform degradations, we present a physical-semantic joint modulation mechanism coupling a continuous blur field with frozen CLIP priors. This emulates focal adaptation by dynamically prioritizing restoration in heavily blurred, semantic regions. Extensive experiments show our method achieves state-of-the-art restoration quality with highly favorable memory and runtime efficiency.


%
%
\bibliographystyle{splncs04}
\bibliography{main}

@String(CVPR  = {IEEE Conf. Comput. Vis. Pattern Recog.})

@String(ICCV  = {Int. Conf. Comput. Vis.})

@String(ECCV  = {Eur. Conf. Comput. Vis.})

@String(AAAI  = {AAAI})

@String(CVPR  = {CVPR})

@String(ICCV  = {ICCV})

@String(ECCV  = {ECCV})

@inproceedings{dong2014learning,
  title={Learning a deep convolutional network for image super-resolution},
  author={Dong, Chao and Loy, Chen Change and He, Kaiming and Tang, Xiaoou},
  booktitle={European conference on computer vision},
  pages={184--199},
  year={2014},
  organization={Springer}
}

@article{zhang2017beyond,
  title={Beyond a gaussian denoiser: Residual learning of deep cnn for image denoising},
  author={Zhang, Kai and Zuo, Wangmeng and Chen, Yunjin and Meng, Deyu and Zhang, Lei},
  journal={IEEE transactions on image processing},
  volume={26},
  number={7},
  pages={3142--3155},
  year={2017},
  publisher={IEEE}
}

@inproceedings{dong2015compression,
  title={Compression artifacts reduction by a deep convolutional network},
  author={Dong, Chao and Deng, Yubin and Loy, Chen Change and Tang, Xiaoou},
  booktitle={Proceedings of the IEEE international conference on computer vision},
  pages={576--584},
  year={2015}
}

@article{gu2021efficiently,
  title={Efficiently modeling long sequences with structured state spaces},
  author={Gu, Albert and Goel, Karan and R{\'e}, Christopher},
  journal={arXiv preprint arXiv:2111.00396},
  year={2021}
}

@article{smith2022simplified,
  title={Simplified state space layers for sequence modeling},
  author={Smith, Jimmy TH and Warrington, Andrew and Linderman, Scott W},
  journal={arXiv preprint arXiv:2208.04933},
  year={2022}
}

@article{fu2022hungry,
  title={Hungry hungry hippos: Towards language modeling with state space models},
  author={Fu, Daniel Y and Dao, Tri and Saab, Khaled K and Thomas, Armin W and Rudra, Atri and R{\'e}, Christopher},
  journal={arXiv preprint arXiv:2212.14052},
  year={2022}
}

@inproceedings{huang2024localmamba,
  title={Localmamba: Visual state space model with windowed selective scan},
  author={Huang, Tao and Pei, Xiaohuan and You, Shan and Wang, Fei and Qian, Chen and Xu, Chang},
  booktitle={European Conference on Computer Vision},
  pages={12--22},
  year={2024},
  organization={Springer}
}

@inproceedings{conde2022swin2sr,
  title={Swin2sr: Swinv2 transformer for compressed image super-resolution and restoration},
  author={Conde, Marcos V and Choi, Ui-Jin and Burchi, Maxime and Timofte, Radu},
  booktitle={European Conference on Computer Vision},
  pages={669--687},
  year={2022},
  organization={Springer}
}

@inproceedings{mprnet,
  title={Multi-stage progressive image restoration},
  author={Zamir, Syed Waqas and Arora, Aditya and Khan, Salman and Hayat, Munawar and Khan, Fahad Shahbaz and Yang, Ming-Hsuan and Shao, Ling},
  booktitle={Proceedings of the IEEE/CVF conference on computer vision and pattern recognition},
  pages={14821--14831},
  year={2021}
}

@inproceedings{nafnet,
  title={Simple baselines for image restoration},
  author={Chen, Liangyu and Chu, Xiaojie and Zhang, Xiangyu and Sun, Jian},
  booktitle={ECCV},
  pages={17--33},
  year={2022},
  organization={Springer}
}

@inproceedings{fftformer,
  title={Efficient frequency domain-based transformers for high-quality image deblurring},
  author={Kong, Lingshun and Dong, Jiangxin and Ge, Jianjun and Li, Mingqiang and Pan, Jinshan},
  booktitle={Proceedings of the IEEE/CVF conference on computer vision and pattern recognition},
  pages={5886--5895},
  year={2023}
}

@inproceedings{airnet,
  title={All-in-one image restoration for unknown corruption},
  author={Li, Boyun and Liu, Xiao and Hu, Peng and Wu, Zhongqin and Lv, Jiancheng and Peng, Xi},
  booktitle={CVPR},
  pages={17452--17462},
  year={2022}
}

@inproceedings{restormer,
  title={Restormer: Efficient transformer for high-resolution image restoration},
  author={Zamir, Syed Waqas and Arora, Aditya and Khan, Salman and Hayat, Munawar and Khan, Fahad Shahbaz and Yang, Ming-Hsuan},
  booktitle={Proceedings of the IEEE/CVF conference on computer vision and pattern recognition},
  pages={5728--5739},
  year={2022}
}

@inproceedings{mambair,
  title={Mambair: A simple baseline for image restoration with state-space model},
  author={Guo, Hang and Li, Jinmin and Dai, Tao and Ouyang, Zhihao and Ren, Xudong and Xia, Shu-Tao},
  booktitle={European conference on computer vision},
  pages={222--241},
  year={2024},
  organization={Springer}
}

@article{vmamba,
  title={Vmamba: Visual state space model},
  author={Liu, Yue and Tian, Yunjie and Zhao, Yuzhong and Yu, Hongtian and Xie, Lingxi and Wang, Yaowei and Ye, Qixiang and Jiao, Jianbin and Liu, Yunfan},
  journal={Advances in neural information processing systems},
  volume={37},
  pages={103031--103063},
  year={2024}
}

@inproceedings{SRN,
  title={Scale-recurrent network for deep image deblurring},
  author={Tao, Xin and Gao, Hongyun and Shen, Xiaoyong and Wang, Jue and Jia, Jiaya},
  booktitle={Proceedings of the IEEE conference on computer vision and pattern recognition},
  pages={8174--8182},
  year={2018}
}

@inproceedings{MIMOUnet,
  title={Rethinking coarse-to-fine approach in single image deblurring},
  author={Cho, Sung-Jin and Ji, Seo-Won and Hong, Jun-Pyo and Jung, Seung-Won and Ko, Sung-Jea},
  booktitle={Proceedings of the IEEE/CVF international conference on computer vision},
  pages={4641--4650},
  year={2021}
}

@inproceedings{MLWNet,
  title={Efficient multi-scale network with learnable discrete wavelet transform for blind motion deblurring},
  author={Gao, Xin and Qiu, Tianheng and Zhang, Xinyu and Bai, Hanlin and Liu, Kang and Huang, Xuan and Wei, Hu and Zhang, Guoying and Liu, Huaping},
  booktitle={Proceedings of the IEEE/CVF Conference on Computer Vision and Pattern Recognition},
  pages={2733--2742},
  year={2024}
}

@inproceedings{evssm,
  title={Efficient visual state space model for image deblurring},
  author={Kong, Lingshun and Dong, Jiangxin and Tang, Jinhui and Yang, Ming-Hsuan and Pan, Jinshan},
  booktitle={Proceedings of the computer vision and pattern recognition conference},
  pages={12710--12719},
  year={2025}
}

@article{banet,
  title={BANet: A blur-aware attention network for dynamic scene deblurring},
  author={Tsai, Fu-Jen and Peng, Yan-Tsung and Tsai, Chung-Chi and Lin, Yen-Yu and Lin, Chia-Wen},
  journal={IEEE Transactions on Image Processing},
  volume={31},
  pages={6789--6799},
  year={2022},
  publisher={IEEE}
}

@inproceedings{algnet,
  title={Learning enriched features via selective state spaces model for efficient image deblurring},
  author={Gao, Hu and Ma, Bowen and Zhang, Ying and Yang, Jingfan and Yang, Jing and Dang, Depeng},
  booktitle={Proceedings of the 32nd ACM International Conference on Multimedia},
  pages={710--718},
  year={2024}
}

@InProceedings{adarev,
    author    = {Mao, Xintian and Li, Qingli and Wang, Yan},
    title     = {AdaRevD: Adaptive Patch Exiting Reversible Decoder Pushes the Limit of Image Deblurring},
    booktitle = {Proceedings of the IEEE/CVF Conference on Computer Vision and Pattern Recognition (CVPR)},
    month     = {June},
    year      = {2024},
    pages     = {25681-25690}
}

@inproceedings{MRLPFNet,
  title={Multi-scale residual low-pass filter network for image deblurring},
  author={Dong, Jiangxin and Pan, Jinshan and Yang, Zhongbao and Tang, Jinhui},
  booktitle={Proceedings of the IEEE/CVF international conference on computer vision},
  pages={12345--12354},
  year={2023}
}

@inproceedings{stripformer,
  title={Stripformer: Strip transformer for fast image deblurring},
  author={Tsai, Fu-Jen and Peng, Yan-Tsung and Lin, Yen-Yu and Tsai, Chung-Chi and Lin, Chia-Wen},
  booktitle={European conference on computer vision},
  pages={146--162},
  year={2022},
  organization={Springer}
}

@article{brdnet,
  title={Image denoising using deep CNN with batch renormalization},
  author={Tian, Chunwei and Xu, Yong and Zuo, Wangmeng},
  journal={Neural Networks},
  volume={121},
  pages={461--473},
  year={2020},
  publisher={Elsevier}
}

@article{lpnet,
  title={Lightweight pyramid networks for image deraining},
  author={Fu, Xueyang and Liang, Borong and Huang, Yue and Ding, Xinghao and Paisley, John},
  journal={IEEE transactions on neural networks and learning systems},
  volume={31},
  number={6},
  pages={1794--1807},
  year={2019},
  publisher={IEEE}
}

@inproceedings{fdgan,
  title={FD-GAN: Generative adversarial networks with fusion-discriminator for single image dehazing},
  author={Dong, Yu and Liu, Yihao and Zhang, He and Chen, Shifeng and Qiao, Yu},
  booktitle={Proceedings of the AAAI conference on artificial intelligence},
  volume={34},
  number={07},
  pages={10729--10736},
  year={2020}
}

@article{DL,
  title={A general decoupled learning framework for parameterized image operators},
  author={Fan, Qingnan and Chen, Dongdong and Yuan, Lu and Hua, Gang and Yu, Nenghai and Chen, Baoquan},
  journal={IEEE transactions on pattern analysis and machine intelligence},
  volume={43},
  number={1},
  pages={33--47},
  year={2019},
  publisher={IEEE}
}

@inproceedings{mspfn,
  title={Multi-scale progressive fusion network for single image deraining},
  author={Jiang, Kui and Wang, Zhongyuan and Yi, Peng and Chen, Chen and Huang, Baojin and Luo, Yimin and Ma, Jiayi and Jiang, Junjun},
  booktitle={Proceedings of the IEEE/CVF conference on computer vision and pattern recognition},
  pages={8346--8355},
  year={2020}
}

@InProceedings{swinir,
    author    = {Liang, Jingyun and Cao, Jiezhang and Sun, Guolei and Zhang, Kai and Van Gool, Luc and Timofte, Radu},
    title     = {SwinIR: Image Restoration Using Swin Transformer},
    booktitle = {Proceedings of the IEEE/CVF International Conference on Computer Vision (ICCV) Workshops},
    month     = {October},
    year      = {2021},
    pages     = {1833-1844}
}

@ARTICLE{idt,
  author={Xiao, Jie and Fu, Xueyang and Liu, Aiping and Wu, Feng and Zha, Zheng-Jun},
  journal={IEEE Transactions on Pattern Analysis and Machine Intelligence}, 
  title={Image De-Raining Transformer}, 
  year={2023},
  volume={45},
  number={11},
  pages={12978-12995},
  doi={10.1109/TPAMI.2022.3183612}}

@inproceedings{gopro,
  title={Deep multi-scale convolutional neural network for dynamic scene deblurring},
  author={Nah, Seungjun and Hyun Kim, Tae and Mu Lee, Kyoung},
  booktitle={Proceedings of the IEEE conference on computer vision and pattern recognition},
  pages={3883--3891},
  year={2017}
}

@inproceedings{hide,
  title={Human-aware motion deblurring},
  author={Shen, Ziyi and Wang, Wenguan and Lu, Xiankai and Shen, Jianbing and Ling, Haibin and Xu, Tingfa and Shao, Ling},
  booktitle={Proceedings of the IEEE/CVF international conference on computer vision},
  pages={5572--5581},
  year={2019}
}

@inproceedings{realblur,
  title={Real-world blur dataset for learning and benchmarking deblurring algorithms},
  author={Rim, Jaesung and Lee, Haeyun and Won, Jucheol and Cho, Sunghyun},
  booktitle={European conference on computer vision},
  pages={184--201},
  year={2020},
  organization={Springer}
}

@article{rain100,
  title={Clearing the skies: A deep network architecture for single-image rain removal},
  author={Fu, Xueyang and Huang, Jiabin and Ding, Xinghao and Liao, Yinghao and Paisley, John},
  journal={IEEE Transactions on Image Processing},
  volume={26},
  number={6},
  pages={2944--2956},
  year={2017},
  publisher={IEEE}
}

@ARTICLE{reside,
  author={Li, Boyi and Ren, Wenqi and Fu, Dengpan and Tao, Dacheng and Feng, Dan and Zeng, Wenjun and Wang, Zhangyang},
  journal={IEEE Transactions on Image Processing}, 
  title={Benchmarking Single-Image Dehazing and Beyond}, 
  year={2019},
  volume={28},
  number={1},
  pages={492-505},
  doi={10.1109/TIP.2018.2867951}}

@inproceedings{bsd68,
  title={A database of human segmented natural images and its application to evaluating segmentation algorithms and measuring ecological statistics},
  author={Martin, David and Fowlkes, Charless and Tal, Doron and Malik, Jitendra},
  booktitle={Proceedings eighth IEEE international conference on computer vision. ICCV 2001},
  volume={2},
  pages={416--423},
  year={2001},
  organization={Ieee}
}
\end{document}